\documentclass[10pt,twocolumn,letterpaper]{article}

\usepackage[pagenumbers]{cvpr} 

\usepackage{url}
\usepackage[utf8]{inputenc} 
\usepackage[T1]{fontenc}    
\usepackage{booktabs}       
\usepackage{amsfonts}       
\usepackage{nicefrac}       
\usepackage{microtype}      
\usepackage{xcolor}         
\usepackage{multirow}
\usepackage{dsfont}
\usepackage{soul}
\usepackage{adjustbox}
\usepackage{xspace}
\usepackage{bm}
\usepackage{bbm}
\usepackage{amsmath}
\usepackage{algorithm}
\usepackage{algpseudocode}                  
\usepackage{setspace}                       
\usepackage{scrextend}
\usepackage{mathtools}
\usepackage{graphicx}
\usepackage{subcaption}
\usepackage{enumitem}
\usepackage{colortbl}
\usepackage{wrapfig}
\usepackage{kotex}
\usepackage{lipsum}
\usepackage{xcolor,pifont}

\usepackage{duckuments}
\usepackage[pagebackref,breaklinks,colorlinks]{hyperref}

\makeatletter
\DeclareRobustCommand\onedot{\futurelet\@let@token\@onedot}
\def\@onedot{\ifx\@let@token.\else.\null\fi\xspace}

\newcommand{\stdv}[1]{\scriptsize$\pm$#1}
\newcommand{\bx}{\mathbf{x}}
\newcommand{\bz}{\mathbf{z}}

\definecolor{pinegreen}{rgb}{0.0, 0.47, 0.44}
\definecolor{cornellred}{rgb}{0.7, 0.11, 0.11}
\definecolor{cadmiumgreen}{rgb}{0.0, 0.42, 0.24}
\definecolor{spirodiscoball}{rgb}{0.06, 0.75, 0.99}
\definecolor{blizzardblue}{rgb}{0.73, 0.96, 0.99}
\definecolor{aliceblue}{rgb}{0.91, 0.94, 0.97}
\definecolor{darkblue}{rgb}{0.83, 0.89, 0.97}
\hypersetup{
  linkcolor = cornellred,
  citecolor  = cadmiumgreen,
  colorlinks = true,
}

\usepackage[capitalize]{cleveref}
\crefname{section}{Sec.}{Secs.}
\Crefname{section}{Section}{Sections}
\Crefname{table}{Table}{Tables}
\crefname{table}{Tab.}{Tabs.}

\def\cvprPaperID{XXXXX} 
\def\confName{CVPR}
\def\confYear{2023}

\newcommand{\lname}{projected latent video diffusion model\xspace} 
\newcommand{\sname}{PVDM\xspace} 

\newcommand{\citep}{\cite}
\newcommand{\citet}{\cite}

\begin{document}

\title{Video Probabilistic Diffusion Models in Projected Latent Space}

\author{Sihyun Yu$^{1}$ \qquad Kihyuk Sohn$^{2}$ \qquad Subin Kim$^{1}$ \qquad Jinwoo Shin$^{1}$\\
$^{1}$KAIST \qquad $^{2}$Google Research\\
{\tt\small \{sihyun.yu, subin-kim, jinwoos\}@kaist.ac.kr, kihyuks@google.com}
}
\maketitle

\begin{abstract}
Despite the remarkable progress in deep generative models, synthesizing high-resolution and temporally coherent videos still remains a challenge due to their high-dimensionality and complex temporal dynamics along with large spatial variations.
Recent works on diffusion models have shown their potential to solve this challenge, yet they suffer from severe computation- and memory-inefficiency that limit the scalability.
To handle this issue, we propose a novel generative model for videos, coined {\lname} (\sname), a probabilistic diffusion model which learns a video distribution in a low-dimensional latent space and thus can be efficiently trained with high-resolution videos under limited resources. 
Specifically, \sname is composed of two components: (a) an autoencoder that projects a given video as 2D-shaped latent vectors that factorize the complex cubic structure of video pixels and (b) a diffusion model architecture specialized for our new factorized latent space and the training/sampling procedure to synthesize videos of arbitrary length with a single model.
Experiments on popular video generation datasets demonstrate the superiority of \sname compared with previous video synthesis methods; e.g., \sname obtains the FVD score of 639.7 on the UCF-101 long video (128 frames) generation benchmark, which improves 1773.4 of the prior state-of-the-art.
\end{abstract}

\section{Introduction}
Recent progresses of deep generative models have shown their promise to synthesize high-quality, realistic samples in various domains, such as images \citep{dhariwal2021diffusion,karras2021alias,rombach2021highresolution}, audio \citep{dhariwal2020jukebox, lakhotia2021generative, kong2020diffwave}, 3D scenes~\citep{chan2022efficient,epigraf,poole2022dreamfusion}, natural languages \citep{adiwardana2020towards,brown2020language}, \emph{etc}. As a next step forward, several works have been actively focusing on the more challenging task of video synthesis~\citep{tian2021good,yu2022digan,skorokhodov2021stylegan,ho2022video,ge2022long,ho2022imagen}. In contrast to the success in other domains, the generation quality is yet far from real-world videos, due to the high-dimensionality and complexity of videos that contain complicated spatiotemporal dynamics in high-resolution frames.

\begin{figure*}[t]
\centering\small
\includegraphics[width=.9\textwidth]{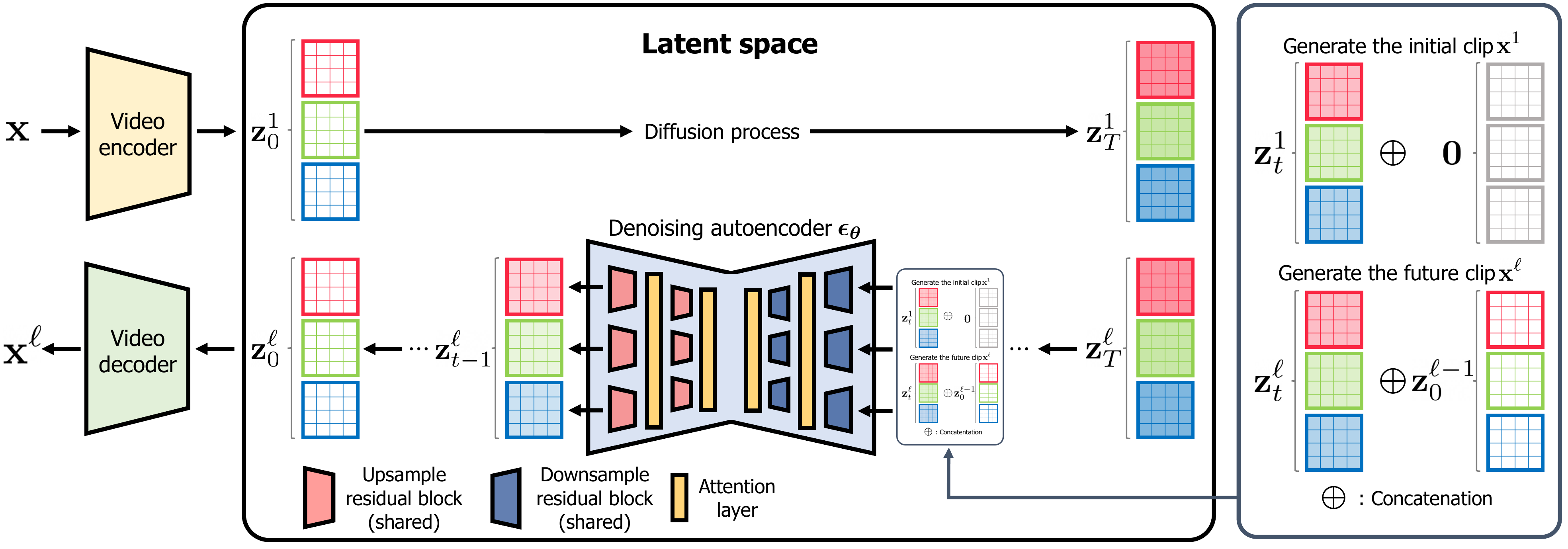}
\caption{
Overall illustration of our \lname (\sname) framework. \sname is composed of two components: (a) (left) an autoencoder that maps a video into 2D image-like latent space (b) (right) a diffusion model operates in this latent space.
}\label{fig:concept}
\end{figure*}

Inspired by the success of diffusion models in handling complex and large-scale image datasets~\citep{dhariwal2021diffusion,ramesh2022hierarchical}, recent approaches have attempted to design diffusion models for videos~\citep{ho2022video,harvey2022flexible,yang2022diffusion,hoppe2022diffusion, ho2022imagen, mei2023vidm}. Similar to image domains, these methods have shown great potential to model video distribution much better with scalability (both in terms of spatial resolution and temporal durations), even achieving photorealistic generation results~\citep{ho2022imagen}. However, they suffer from severe computation and memory inefficiency, as diffusion models require lots of iterative processes in input space to synthesize samples~\citep{song2021denoising}. Such bottlenecks are much more amplified in video due to a cubic RGB array structure. 

Meanwhile, recent works in image generation have proposed latent diffusion models to circumvent the computation and memory inefficiency of diffusion models \citep{vahdat2021score, rombach2021highresolution,gu2021vector}. Instead of training the model in raw pixels, latent diffusion models first train an autoencoder to learn a low-dimensional latent space succinctly parameterizing images~\citep{van2017neural,esser2020taming,rombach2021highresolution} and then models this latent distribution. Intriguingly, the approach has shown a dramatic improvement in efficiency for synthesizing samples while even achieving state-of-the-art generation results~\citep{rombach2021highresolution}. Despite their appealing potential, however, developing a form of latent diffusion model for videos is yet overlooked.

\vspace{0.02in}
\noindent\textbf{Contribution.}
We present a novel latent diffusion model for videos, coined \emph{\lname} (\sname). Specifically, it is a two-stage framework (see Figure~\ref{fig:concept} for the overall illustration):
\begin{itemize}[topsep=0.0pt,itemsep=1.2pt,leftmargin=5.5mm]
    \item \emph{Autoencoder:}
    We introduce an autoencoder that represents a video with \emph{three 2D image-like latent vectors} by factorizing the complex cubic array structure of videos. Specifically, we propose 3D $\rightarrow$ 2D projections of videos at each spatiotemporal direction to encode 3D video pixels as three succinct 2D latent vectors. At a high level, we design one latent vector across the temporal direction to parameterize the common contents of the video (\eg, background), and the latter two vectors to encode the motion of a video. These 2D latent vectors are beneficial for achieving high-quality and succinct encoding of videos, as well as enabling compute-efficient diffusion model architecture design due to their image-like structure.

    \item \emph{Diffusion model:} 
    Based on the 2D image-like latent space built from our video autoencoder, we design a new diffusion model architecture to model the video distribution. Since we parameterize videos as image-like latent representations, we avoid computation-heavy 3D convolutional neural network architectures that are conventionally used for handling videos. Instead, our architecture is based on 2D convolution network diffusion model architecture that has shown its strength in handling images. Moreover, we present a joint training of unconditional and frame conditional generative modeling to generate a long video of arbitrary lengths.
\end{itemize}

\vspace{0.02in}
We verify the effectiveness of our method on two popular datasets for evaluating video generation methods: UCF-101 \citep{soomro2012ucf101} and SkyTimelapse \citep{xiong2018learning}.
Measured with Inception score (IS; higher is better \citep{salimans2016improved}) on UCF-101, a representative metric of evaluating unconditional video generation, \sname achieves the state-of-the-art result of 74.40 on UCF-101 in generating 16 frames, 256$\times$256 resolution videos. In terms of Fr\'echet video distance (FVD; lower is better~\citep{unterthiner2018towards}) on UCF-101 in synthesizing long videos (128 frames) of 256$\times$256 resolution, it significantly improves the score from 1773.4 of the prior state-of-the-art to 639.7. Moreover, compared with recent video diffusion models, our model shows a strong memory and computation efficiency. For instance, on a single NVIDIA 3090Ti 24GB GPU, a video diffusion model~\cite{ho2022video} requires almost full memory ($\approx$24GB) to train at 128$\times$128 resolution with a batch size of 1. On the other hand, \sname can be trained with a batch size of 7 at most per this GPU with 16 frames videos at 256${\times}$256 resolution.

To our knowledge, the proposed \sname is the first latent diffusion model designed for video synthesis. We believe our work would facilitate video generation research towards efficient real-time, high-resolution, and long video synthesis under the limited computational resource constraints. 

\begin{figure*}[t!]
\centering\small
\includegraphics[width=.9\textwidth]{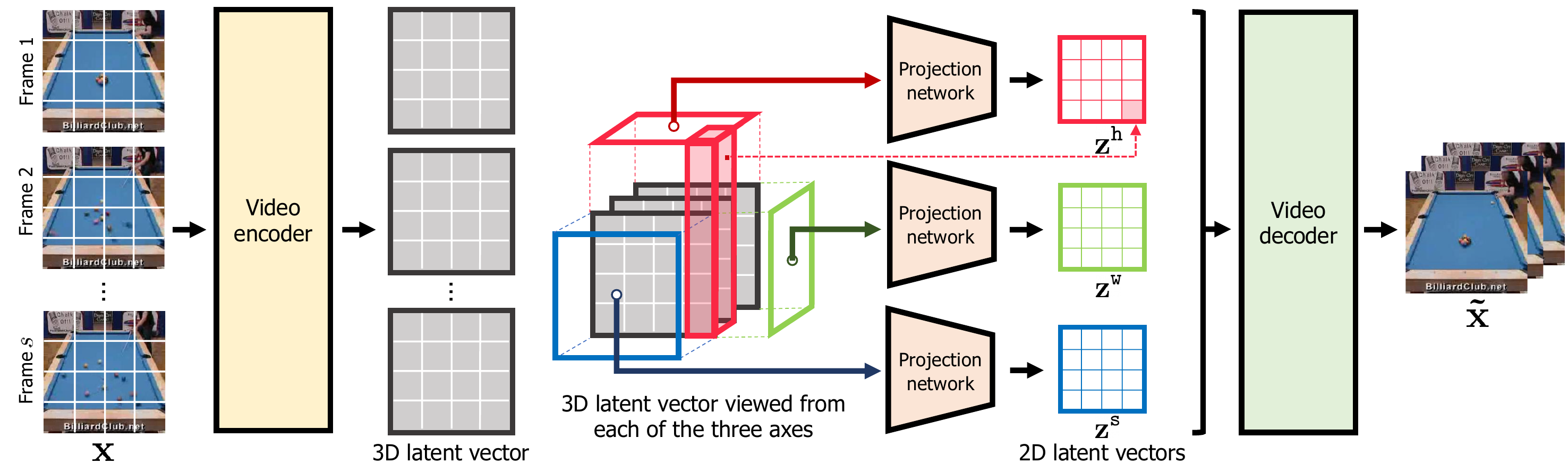}
\caption{
Detailed illustration of our autoencoder architecture in \sname framework ((a) in Figure~\ref{fig:concept}). }\label{fig:autoencoder}
\vspace{-0.1in}
\end{figure*}

\section{Related work}
\noindent\textbf{Video generation.}
Video generation is one of the long-standing goals in deep generative models. Many prior works have attempted to solve the problem and they mostly fall into three categories. First, there exists numerous attempts to extend image generative adversarial networks (GANs) \citep{goodfellow2014generative} to generate videos~\citep{vondrick2016generating,saito2017temporal,saito2020train,tulyakov2018mocogan,acharya2018towards,clark2019adversarial,yushchenko2019markov,kahembwe2020lower,gordon2021latent,tian2021good, fox2021stylevideogan,munoz2021temporal,yu2022digan,skorokhodov2021stylegan,singer2022make}; however, GANs often suffer from mode collapse problem and these methods are difficult to be scaled to complex, large-scale video datasets. Other approaches have proposed learning the distribution via training autoregressive models \citep{kalchbrenner2017video,weissenborn2020scaling,rakhimov2020latent,yan2021videogpt,ge2022long} using Transformers~\citep{vaswani2017attention}. They have shown better mode coverage and video quality than GAN-based approaches, but they require expensive computation and memory costs to generate longer videos~\citep{skorokhodov2021stylegan}. Finally, recent works have attempted to build diffusion models~\citep{ho2021denoising} for videos~\citep{ho2022video,harvey2022flexible,yang2022diffusion,hoppe2022diffusion,singer2022make}, achieving state-of-the-art results, yet they also suffer from significant computation and memory inefficiency. Our method also takes an approach to diffusion models for videos; however, we consider the generative modeling in low-dimensional latent space to alleviate these bottlenecks of diffusion models.

\vspace{0.02in}
\noindent\textbf{Diffusion models.}
Diffusion models \citep{sohl2015deep,ho2021denoising}, which are categorized as score-based generative models \citep{song2019generative,song2021scorebased}, model the data distribution by learning a gradual iterative denoising process from the Gaussian distribution to the data distribution. Intriguingly, they show a strong promise in generating high-quality samples with wide mode coverage, even outperforming GANs in image synthesis \citep{dhariwal2021diffusion} and enabling zero-shot text-to-image synthesis~\citep{ramesh2022hierarchical}. Not limited to images, diffusion models have shown their promise in other data domains, including point clouds \citep{luo2021diffusion}, audio \citep{kong2020diffwave}, etc. However, diffusion models suffer from severe computation inefficiency for data sampling due to the iterative denoising process in high-dimensional data space. To alleviate this problem, several works have proposed an effective sampling process from trained diffusion models \citep{song2021denoising,zhang2022fast} or learning the data distribution from low-dimensional latent space that amortizes the data \citep{vahdat2021score, rombach2021highresolution,gu2021vector}. We take the latter approach for designing a computation-efficient video diffusion model.

\vspace{0.02in}
\noindent\textbf{Diffusion models for videos.} Following the remarkable success of diffusion models in image domains, several works \citep{ho2022video,harvey2022flexible,yang2022diffusion,hoppe2022diffusion,ho2022imagen,mei2023vidm} have extended them for video generation. Intriguingly, they often show much better results than prior works, even can be scaled-up to complex datasets, and achieves successful generation results on challenging text-to-video 
synthesis task~\citep{ho2022imagen}. Despite their potential on modeling videos, scaling them for synthesizing high-resolution, long video is not straightforward due to a huge computation bottleneck caused by an unfavorable dimension increase of video data as 3D RGB arrays. We tackle this problem by modeling the video distribution in the low-dimensional latent space.

\vspace{0.02in}
\noindent\textbf{Triplane representations.}
Several recent works in 3D-aware generation~\citep{chan2022efficient,epigraf} have demonstrated that high-dimensional 3D voxels can be effectively parameterized with 2D triplane latent representations without sacrificing the encoding quality. In particular, prior works have proposed GAN architectures that synthesize a 3D scene by generating three image-like latents as an approximation of complex 3D voxels. In contrast, we explore the effectiveness of triplane representations for encoding videos, and we build diffusion models instead of GANs with such representations.

\vspace{-0.07in}
\section{Projected latent video diffusion model}
We first formalize the problem of \emph{video generative modeling}. Consider a dataset $\mathcal{D} \,{=}\, \{\bx_i\}_{i=1}^{N}$ of size $N$, where each $\bx \,{\in}\, \mathcal{D}$ is sampled from the unknown data distribution $p_{\text{data}}(\bx)$. Here, each $\bx \in \mathbb{R}^{3 \times S \times H \times W}$ is a \emph{video clip} consisting of $S$ frames at $H \times W$ (spatial) resolution. In video generative modeling, the goal is to learn a model distribution $p_{\text{model}}(\bx)$ that matches $p_{\text{data}}(\bx)$ using $\mathcal{D}$.

To accomplish the goal, we build a method based on diffusion models---a type of generative model that models the data distribution $p_{\text{data}}(\bx)$ by learning the reverse process of the Markov diffusion process starting from $p_{\text{data}}(\bx)$ to the Gaussian prior distribution $\mathcal{N}(\mathbf{0}_\bx, \mathbf{I}_\bx)$. Remarkably, diffusion models are shown to synthesize high-quality samples without mode collapse problems and can be scaled to model complex datasets. However, diffusion models directly operate in raw input space; unlike other data domains (\eg, images), designing diffusion models for videos is challenging due to their cubic complexity and high-dimensionality as 3D tensors of RGB values. Our key contribution is to mitigate this issue by proposing diffusion models operating on a novel low-dimensional latent space that succinctly parameterizes videos by breaking down the complex 3D structure of video pixels into three 2D structures.

In the rest of this section, we describe our \emph{\lname} (\sname) in detail. In Section~\ref{subsec:ldm}, we provide an overview of (latent) diffusion models. In Section~\ref{subsec:ours}, we explain how we design \sname in detail. Finally, in Section~\ref{sucsec:long}, we describe our training objective and sampling strategy to generate longer videos.

\subsection{Latent diffusion models}
\label{subsec:ldm}
At a high level, diffusion models learn the target distribution $p_{\text{data}}(\bx)$ by learning a gradual denoising process from Gaussian prior distribution to reach $p_{\text{data}}(\bx)$. Formally, diffusion models consider the reverse process $p_{\bm{\theta}}(\bx_{t-1}|\bx_{t})$ of the Markov diffusion process $q(\bx_{t}|\bx_{t-1})$ of a fixed length $T>0$ starting from $p(\bx_0)\,{\coloneqq}\, p_{\text{data}}(\bx)$ to $p(\bx_T) \,{\coloneqq}\,\mathcal{N}(\mathbf{0}_{\bx}, \mathbf{I}_{\bx})$. More specifically, $q(\bx_{t}|\bx_{t-1})$ is formalized as the following normal distribution with a pre-defined $0<\beta_1,\ldots,\beta_T<1$ and $\bar{\alpha}_t \coloneqq \prod_{i=1}^{t} (1-\beta_i)$:
\begin{align*}
    q(\bx_{t}|\bx_{t-1}) &\coloneqq \mathcal{N}(\bx_t; \sqrt{1-\beta_t}\bx_{t-1}, \beta_t\mathbf{I}_\bx), \\ q(\bx_t| \bx_0) &= \mathcal{N}(\bx_t; \sqrt{\bar{\alpha}_t}\bx_0, (1-\bar{\alpha}_t)\mathbf{I}_\bx).
\end{align*}
Using the reparameterization trick \citep{kingma2013auto}, Ho et al.\citet{ho2021denoising} shows the corresponding $p_{\bm{\theta}}(\bx_{t-1}|\bx_{t})$ can be learned as a denoising autoencoder $\bm{\epsilon}_{\bm{\theta}}(\bx_t, t)$ that denoises a noisy sample $\bx_{t}$, which is trained with the following noise-prediction objective:
\begin{align*}
    \mathbb{E}_{\bx_0, \bm{\epsilon}, t} \Big[ ||\bm{\epsilon} - \bm{\epsilon}_{\bm{\theta}}(\bx_t, t)||_2^2 \Big]\,\, \text{where } \bx_t = \sqrt{\bar{\alpha}_t}\bx_0 + \sqrt{1-\bar{\alpha}_t}\bm{\epsilon},
\end{align*}
and $p_{\bm{\theta}}(\bx_{t-1}|\bx_{t})$ can be approximately formulated as the following normal distribution with a small enough $\beta_t$~\citep{pmlr-v37-sohl-dickstein15}:
\begin{align*}
    p_{\bm{\theta}}(\bx_{t-1}|\bx_{t}) \coloneqq \mathcal{N}\Big(\bx_{t-1}; \bx_t - \frac{\beta_t}{\sqrt{1-\bar{\alpha}_t}}\bm{\epsilon}_{\bm{\theta}}(\bx_t, t), \sigma_t^2\Big),
\end{align*}
with the variances $\sigma_t^2 \coloneqq \beta_t$ as pre-defined hyperparameters.

The main drawback of diffusion models is severe computation and memory inefficiency. To generate the sample, one should operate $p_{\bm{\theta}}(\bx_{t-1}|\bx_{t})$ in high-dimensional input space $\mathcal{X}$ repeatedly (\eg, $T\,{=}\,1000$ in Ho et al. \citep{ho2021denoising}). To tackle this issue, several works \citep{vahdat2021score, rombach2021highresolution,gu2021vector} have proposed latent diffusion models to learn the distribution in low-dimensional latent space $\mathcal{Z}$ that succinctly encodes the data, which is typically learned with autoencoders~\citep{rombach2021highresolution}. Specifically, latent diffusion models train the denoising autoencoder $\bm{\epsilon}_{\bm{\theta}}(\bz_t, t)$ in $\mathcal{Z}$ instead of $\mathcal{X}$ \ie, learning $p_{\bm{\theta}}(\bz_{t-1}|\bz_{t})$, so that $\bx$ is generated by first sampling $\bz$ and then decoding to $\bx$ with the decoder. Due to the significant dimension reduction from $\bx$ to $\bz$, one can dramatically reduce the computation for sampling the data. Inspired by their success, we also model video distribution with latent diffusion models.

\begin{algorithm}[t]
\begin{spacing}{1.05}
\caption{\lname (\sname)}\label{algo:1}
\begin{algorithmic}[1]
\For{$\ell=1$ to $L$} \Comment{{\it Iteratively generate video clips $\bx^\ell$.}}
\State Sample the random noise $\bz_T^{\ell} \sim p(\bz_T)$.
\For{$t=T$ to $1$}
\If{$\ell=1$}
\State Unconditional score $\bm{\epsilon}_t = \bm{\epsilon}_{\bm{\theta}}(\bz_t^{\ell}, \mathbf{0}, t)$.
\Else
\State Conditional score $\bm{\epsilon}_t = \bm{\epsilon}_{\bm{\theta}}(\bz_t^{\ell}, \bz_0^{\ell-1}, t)$.
\EndIf
\State Sample $\bm{\epsilon}\sim\mathcal{N}(\mathbf{0}_{\bz}, \mathbf{I}_{\bz})$.
\State Compute $\bz_{t-1}^{\ell}=\frac{1}{\sqrt{1-\beta_t}}\Big(\bz_t^{\ell} - \frac{\beta_t}{\sqrt{1-\bar{\alpha}_t}}\bm{\epsilon}_t \Big) +\sigma_t \bm{\epsilon}$.
\EndFor
\State Decode the $\ell$-th clip $\bx^{\ell} = g_{\bm{\psi}}(\bz_0^{\ell})$.
\EndFor
\State Output the generated video $[\bx^1,\ldots,\bx^{L}]$.
\end{algorithmic}
\end{spacing}
\end{algorithm}

\begin{figure*}[ht!]
\centering\small
\begin{subfigure}{\textwidth}
\centering
\includegraphics[width=\textwidth]{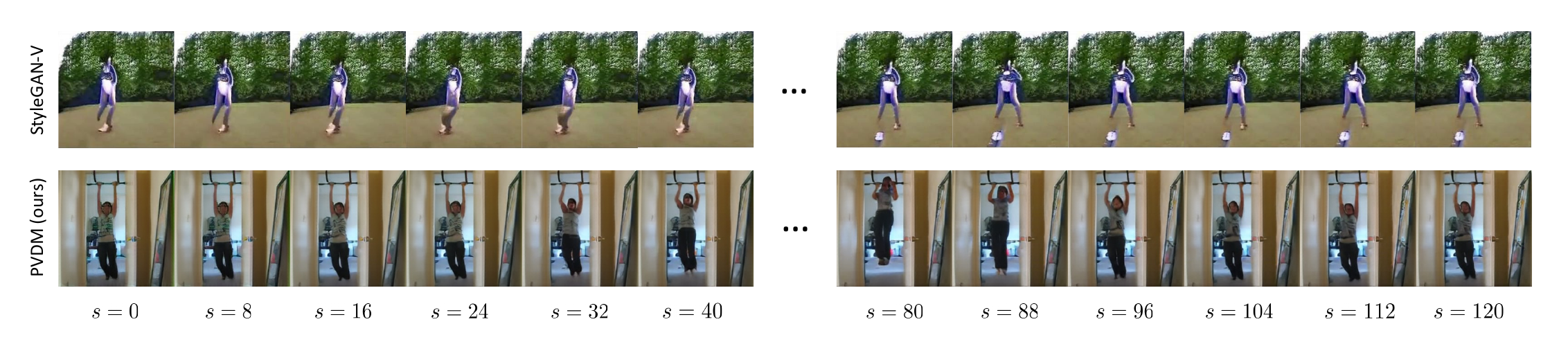}
\vspace{-0.03in}
\caption{UCF-101} 
\vspace{-0.03in}
\label{subfig:emph_ucf}
\end{subfigure}
\begin{subfigure}{\textwidth}
\centering
\includegraphics[width=\textwidth]{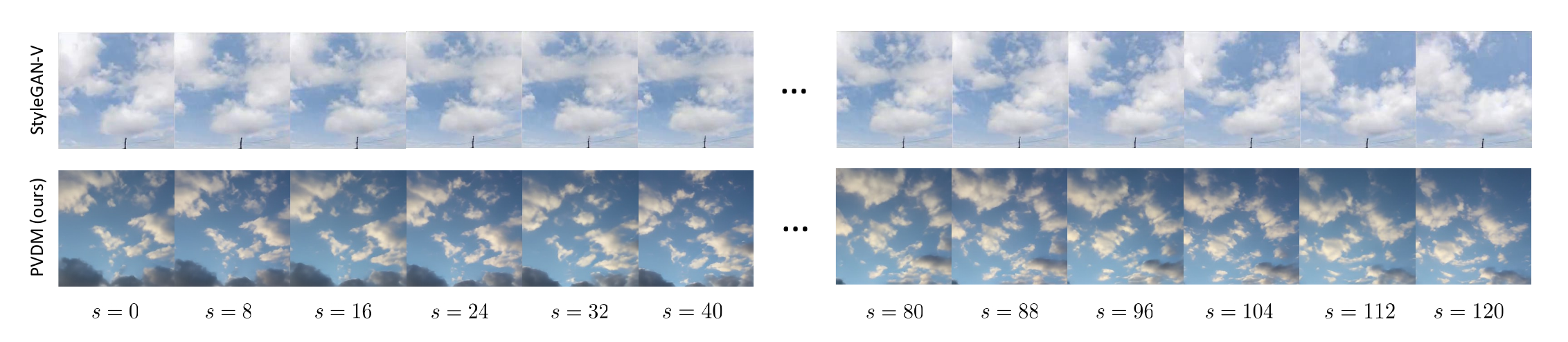}
\vspace{-0.03in}
\caption{SkyTimelapse} 
\label{subfig:emph_sky}
\end{subfigure}
\caption{
256$\times$256 resolution, 128 frame video synthesis results of StyleGAN-V and \sname, trained on (a) UCF-101 and (b) SkyTimelapse.\protect\footnotemark}
\label{fig:emph}
\vspace{-0.05in}
\end{figure*}

\subsection{Designing efficient latent video diffusion model}
\label{subsec:ours}
\noindent\textbf{Autoencoder.}
To represent a video $\bx$ as a low-dimensional latent vector $\bz$, we train an autoencoder composed of an encoder $f_{\bm{\phi}}: \mathcal{X} \rightarrow \mathcal{Z}$ with $f_{\bm{\phi}}(\bx) = \bz$ and a decoder $g_{\bm{\psi}}: \mathcal{Z} \rightarrow \mathcal{X}$ with $g_{\bm{\psi}}(\bz) = \tilde{\bx}$ so that $\tilde{\bx}$ becomes $\bx$. Motivated by VQGAN~\citep{esser2020taming} for compressing images perceptually via latent codes~\citep{oord2018parallel}, we encode videos with discrete latent codes by minimizing the sum of two terms: pixel-level reconstruction loss and the negative of perceptual similarity (LPIPS; \citep{zhang2018perceptual}); we provide more details 
in Appendix~\ref{appen:objective}. 

Conventional video autoencoders have mostly relied on frame-wise 2D convolutional networks \citep{seo2022autoregressive} or 3D convolutional networks~\citep{yan2021videogpt} to compress a given video $\bx$. While these approaches are fairly effective in amortizing $\bx$ as a low-dimensional latent vector, they encode $\bx$ as a 3D latent vector, which overlooks the temporal coherency and requiring the diffusion model architecture to deal with a 3D tensor and may cause a computation overhead. Instead, we take a different approach: given $\bx$, we construct $\bz$ as three 2D latent vectors $\bz^{\tt s}, \bz^{\tt h}, \bz^{\tt w}$, \ie, $\bz \coloneqq[ \bz^{\tt s}, \bz^{\tt h}, \bz^{\tt w}]$, $\bz^{\tt s} \in \mathbb{R}^{C \times H' \times W'}$, $\bz^{\tt h} \in \mathbb{R}^{C \times S \times W'}$, $\bz^{\tt w} \in \mathbb{R}^{C \times S \times H'}$, where $C$ is a latent dimension and $H'=H/d$, $W'=W/d$ for $d>1$. Here, we denote each $\bz^{\tt s}, \bz^{\tt h}, \bz^{\tt w}$ the concatenation of latent codes $z_{hw}^{\tt s}, z_{sw}^{\tt h}, z_{sh}^{\tt w}\in \mathbb{R}^C$ (respectively); \eg, for $\bz^{\tt s}$, we write
\begin{align*}
    \bz^{\tt s} \coloneqq [z_{hw}^{\tt s}]\quad \text{for}\,\, 1\leq h \leq H', 1 \leq w \leq W'.
\end{align*}

We design $\bz^{\tt s}$ to capture the common content across time in $\bx$ (\eg, background), and the latter two latent vectors $\bz^{\tt h}, \bz^{\tt w}$ to encode the underlying motion in $\bx$ by learning the representations across two spatial axes of videos. Specifically, $[\bz^{\tt s}, \bz^{\tt h}, \bz^{\tt w}]$ is computed with the encoder $f_{\bm{\phi}}$, where $f_{\bm{\phi}}$ is a composition of a video-to-3D-latent mapping $f_{\bm{\phi}_{\tt shw}}^{\tt shw}$ and a 3D-to-2D-latents projection $f_{\bm{\phi}_{\tt s}}^{\tt s} \times f_{\bm{\phi}_{\tt h}}^{\tt h} \times f_{\bm{\phi}_{\tt w}}^{\tt w}$ (with $\bm{\phi}\coloneqq (\bm{\phi}_{\tt shw}, \bm{\phi}_{\tt s}, \bm{\phi}_{\tt h}, \bm{\phi}_{\tt w})$; see Figure~\ref{fig:autoencoder} for the illustration). More specifically, we compute $\bz$ from $\bx$ as follows:
\begin{align*}
    \mathbf{u} &\,{\coloneqq}\, f_{\bm{\phi}_{\tt shw}}^{\tt shw}(\bx), \quad \text{where}\,\, \mathbf{u}=[u_{shw}]\in \mathbb{R}^{C\times S \times H' \times W'}, \\ 
    z_{hw}^{\tt s} &\,{\coloneqq}\, f_{\bm{\phi}_{\tt s}}^{\tt s}(u_{1hw},\ldots,u_{Shw}),\,\, 1\,{\leq}\, h \,{\leq}\, H', \,\, 1 \,{\leq}\, w \,{\leq}\, W', \\
    z_{sw}^{\tt h} &\,{\coloneqq}\, f_{\bm{\phi}_{\tt h}}^{\tt h}(u_{s1w},\ldots,u_{sH'w}),\,\, 1\,{\leq}\, s \,{\leq}\, S, \,\, 1 \,{\leq}\, w \,{\leq}\, W', \\
    z_{sh}^{\tt w} &\,{\coloneqq}\, f_{\bm{\phi}_{\tt w}}^{\tt w}(u_{sh1},\ldots,u_{shW'}),\,\, 1\,{\leq}\, s \,{\leq}\, S, \,\, 1 \,{\leq}\, h \,{\leq}\, H'.
\end{align*}

With the latent vector $\bz$ of $\bx$ from $f_{\bm{\phi}}$, we construct the decoder $g_{\bm{\psi}}$ that (a) computes a 3D latent grid $\mathbf{v}$ from $\bz$ and (b) reconstructs $\bx$ from $\mathbf{v}$, where $\mathbf{v}$ is computed as follows:
\begin{gather*}
    \mathbf{v}=(v_{shw})\in\mathbb{R}^{3C\times S \times H' \times W'}  ,\,\,
    v_{shw}\coloneqq[z_{hw}, z_{sw}, z_{sh}].
\end{gather*}

We use video Transformer (\eg, TimeSformer~\citep{bertasius2021space}) for $f_{\bm{\phi}_{\tt shw}}^{\tt shw}(\bx)$ and $g_{\bm{\psi}}$, and a small Transformer~\citep{vaswani2017attention} for projections $f_{\bm{\phi}_{\tt s}}^{\tt s}, f_{\bm{\phi}_{\tt h}}^{\tt h}, f_{\bm{\phi}_{\tt w}}^{\tt w}$. While our autoencoder design requires slightly more parameters and computations for encoding videos from additional projections, it provides dramatic computation efficiency for training (and sampling) of a diffusion model in latent space. In particular, conventional video autoencoder design requires $O(SHW)$ dimensions of latent codes for encoding videos, and thus diffusion models to utilize 3D convolution layers and self-attention layers, which lead the computation overhead to be $O(SHW)$ and $O((SHW)^2)$, respectively. In contrast, we represent a video as image-like latent vectors with $O(HW\,{+}\,SW\,{+}\,SH)$ latent codes; such latent representations enables more compute-efficient diffusion model design by utilizing 2D convolution layers ($O(HW \,{+}\,SW \,{+}\,SH)$) and self-attention layers ($O((HW \,{+}\, SW \,{+}\, SH)^2)$) based on modifying popular architectures used for image diffusion models~\citep{ho2021denoising}.

The intuition behind our overall autoencoder design is that videos are temporally coherent signals and share the common contents across the temporal axis; capturing the common content as $\bz^{\tt s}$ can dramatically reduce the number of parameters for encoding videos \citep{kim2022scalable}. Moreover, due to the high temporal coherency of videos, the temporal variation is often not very large; we empirically verify our succinct representation of motions as two spatial grids $\bz^{\tt h}, \bz^{\tt w}$ does not hurt the encoding quality (see Section~\ref{subsec:exp_ablation}).

\vspace{0.02in}
\noindent\textbf{Latent diffusion model.}
Recall that we encode the video $\bx$ as three 2D latent vectors $\bz = [\bz^{\tt s}, \bz^{\tt h}, \bz^{\tt w}]$; it is enough to model the distribution $p(\bz^{\tt s}, \bz^{\tt h}, \bz^{\tt w})$ for learning $p_{\text{data}} (\bx)$. To train a denoising autoencoder for $[\bz^{\tt s}, \bz^{\tt h}, \bz^{\tt w}]$, we design the neural network architecture based on utilizing popular 2D convolutional U-Net architecture used for training diffusion models for image generation~\citep{dhariwal2021diffusion} instead of 3D convolutional networks~\citep{ho2022video}. Specifically, we use a single U-Net (\ie, shared parameters) to denoise each $\bz^{\tt s}, \bz^{\tt h}, \bz^{\tt w}$. To handle the dependency among $\bz^{\tt s}, \bz^{\tt h}, \bz^{\tt w}$ to model the joint distribution $p(\bz^{\tt s}, \bz^{\tt h}, \bz^{\tt w})$, we add attention layers that operates to the intermediate features of $\bz^{\tt s}, \bz^{\tt h}, \bz^{\tt w}$ from shared U-Net. We remark that such 2D convolutional architecture design is more computation-efficient than a na\"ive 3D convolutional U-Nets for videos, which is possible due to the ``image-like'' structure and a reduced dimension of our latent vectors from using less latent codes for encoding videos.\footnotetext{StyleGAN-V results are from \url{https://universome.github.io/stylegan-v}.}

\subsection{Generating longer videos with \sname}
\label{sucsec:long}
Note that videos are sequential data; unlike our setup that assumes all videos $\bx \in \mathcal{D}$ have the same length $S$, the length of real-world videos varies, and generative video models should be able to generate videos of arbitrary length. However, we only learn the distribution of the fixed-length video clips $p_{\text{data}}(\bx)$; to enable long video generation, one can consider learning a conditional distribution $p(\bx^{2} | \bx^{1})$ of two consecutive video clips $[\bx^{1}, \bx^{2}]$ of $\bx^{1}, \bx^{2} \in \mathbb{R}^{3 \times S \times H \times W}$ and sequentially generate future clips given the current ones.

\begin{figure*}[t!]
\centering\small
\begin{subfigure}{\textwidth}
\centering
\includegraphics[width=\textwidth]{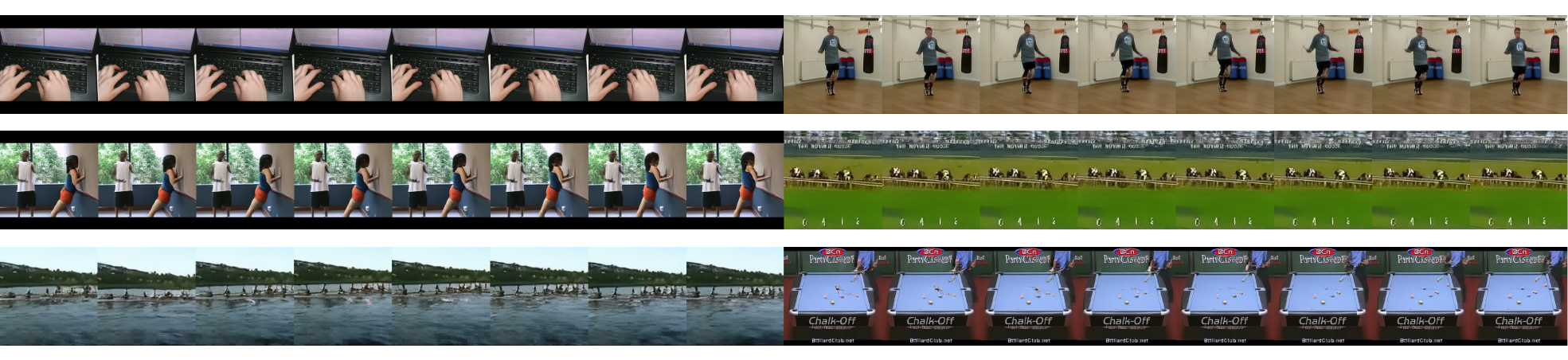}
\caption{UCF-101} 
\label{subfig:ucf}
\end{subfigure}
\begin{subfigure}{\textwidth}
\centering
\includegraphics[width=\textwidth]{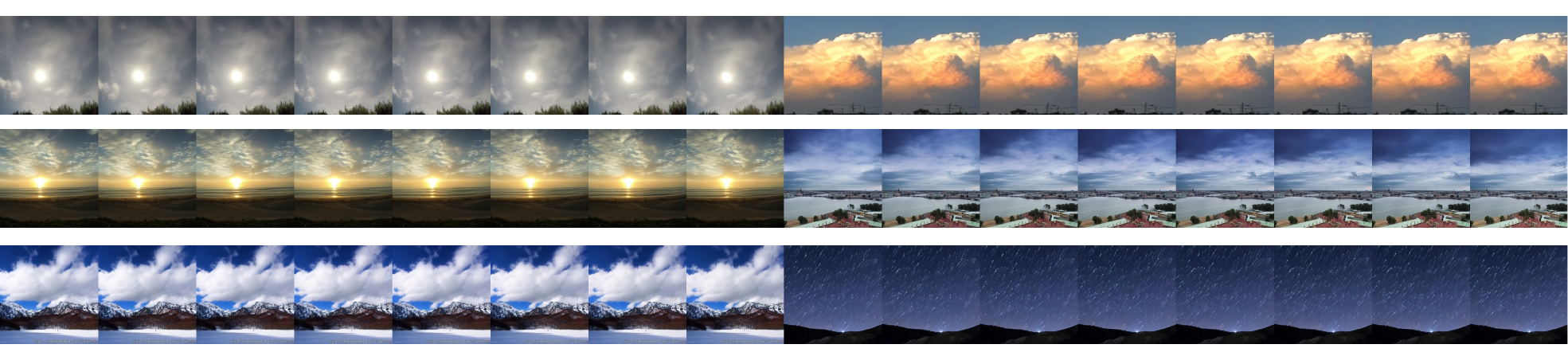}
\caption{SkyTimelapse} 
\label{subfig:sky}
\end{subfigure}
\caption{
Illustrations of random 16 frames, 256$\times$256 resolution video synthesis results of \sname trained on UCF-101 and SkyTimelapse datasets. We visualize the frames of each video with stride 2.
}
\label{fig:main}
\end{figure*}

A straightforward solution is to have two separate models to learn the unconditional distribution $p_{\text{data}}(\bx)$ and the conditional distribution $p(\bx^{2} | \bx^{1})$. Instead of having an extra model, we propose to train a \emph{single} diffusion model to jointly learn an unconditional distribution $p(\bx)$ and the conditional distribution $p(\bx^{2} | \bx^{1})$. It can be achieved by training a conditional diffusion model $p(\bx^{2} | \bx^{1})$ with introducing a null frame (\ie, $\bx^{1}=\bm{0}$) for a joint learning of $p(\bx)$ \citep{ho2022classifier}. More specifically, we consider training of a denoising autoencoder $\bm{\epsilon}_{\bm{\theta}}(\bz_t^{2}, \bz_0^{1}, t)$ in the latent space with the following objective:
\begin{align*}
\vspace{-0.02in}
    \mathbb{E}_{(\bx_0^{1}, \bx_0^{2}), \bm{\epsilon}, t} 
    \Big[ &\lambda||\bm{\epsilon} - \bm{\epsilon}_{\bm{\theta}}(\bz_t^2, \bz_0^1, t)||_2^2 \\ 
    &+ (1-\lambda)||\bm{\epsilon} - \bm{\epsilon}_{\bm{\theta}}(\bz_t^2, \bm{0}, t)||_2^2 
    \Big],
\vspace{-0.02in}
\end{align*}
where $\bz_0^1\,{=}\,f_{\bm{\phi}}({\bx_0^1})$, $\bz_0^2\,{=}\,f_{\bm{\phi}}({\bx_0^2})$, $\bz_t^2 \,{=}\, \sqrt{\bar{\alpha}_t}\bz_0^2 \,{+}\, \sqrt{1-\bar{\alpha}}_t\bm{\epsilon}$, and $\lambda \,{\in}\, (0, 1)$ is a hyperparameter that balances a learning between unconditional and conditional distribution. 

After training, one can generate the long video as follows: sample an initial video clip $\bx^1\,{\sim}\, p_{\bm{\theta}} (\bx)$, and repeat generating next clip $\bx^{\ell+1}\,{\sim}\, p_{\bm{\theta}} (\bx^{\ell+1} |\bx^{\ell})$ conditioned on the previous clip. Finally, we obtain a long video by concatenating all generated clips $[\bx^{1},\ldots,\bx^{ L}]$ of arbitrary length $L>1$. See Algorithm~\ref{algo:1} for details and Figure~\ref{fig:emph} for results.

\section{Experiments}

We validate the superiority of our \sname framework under two representative datasets: UCF-101~\citep{soomro2012ucf101} and SkyTimelapse~\citep{xiong2018learning}. In Section~\ref{subsec:exp_setup}, we provide the experimental setup and evaluation details that we used for experiments. In Section~\ref{subsec:exp_main}, we present the main qualitative and quantitative video synthesis results. Finally, in Section~\ref{subsec:exp_ablation}, we perform an ablation studies to verify the effect of components and efficiencies of our method.

\subsection{Experiment setup}
\label{subsec:exp_setup}
\noindent\textbf{Datasets.} 
We train our \sname and compare with baselines on well-known video datasets used for video synthesis: UCF-101~\citep{soomro2012ucf101} and SkyTimelapse~\citep{xiong2018learning}. Following the experimental setup used in recent video generation methods~\citep{skorokhodov2021stylegan,yu2022digan}, we preprocess these datasets as video clips of length 16 or 128 frames, where each frame is resized to 256$\times$256 resolution. For training each model, we use the train split of the dataset for all of the experiments. We provide the detailed description of datasets in Appendix~\ref{appen:dataset}.

\begin{figure*}[t]
\begin{minipage}[hbt!]{.64\textwidth}
\captionof{table}{
FVD$_{16}$ and FVD$_{128}$ values (lower values are better) of video generation models on UCF-101 and SkyTimelapse. Bolds indicate the best results, and we mark our method as blue. We report FVD values of other baselines obtained by the reference (StyleGAN-V~\citep{skorokhodov2021stylegan}). 
$N$/$M$-s denotes the model is evaluated with the DDIM sampler~\citep{song2021denoising} with $N$ steps (for the initial clip) and $M$ steps (for future clips).
}\label{tab:main} 
\centering\small
\vspace{0.05in}
\begin{tabular}{l c c c c}
\toprule
& \multicolumn{2}{c}{UCF-101}
& \multicolumn{2}{c}{SkyTimelapse}  
\\
\cmidrule(lr){2-3}  \cmidrule(lr){4-5}  
Method 
& {$\text{FVD}_{16}$} $\downarrow$& {$\text{FVD}_{128}$ $\downarrow$} 
& {$\text{FVD}_{16}$ $\downarrow$} & {$\text{FVD}_{128}$ $\downarrow$} 
\\
\midrule
    VideoGPT~\citep{yan2021videogpt}        
    & 2880.6 & N/A    & 222.7 & N/A  
    \\
    MoCoGAN~\citep{tulyakov2018mocogan}         
    & 2886.8 & 3679.0 & 206.6 & 575.9 \\
    + StyleGAN2~\citep{karras2020analyzing}     
    & 1821.4 & 2311.3 & 85.88 & 272.8 \\
    MoCoGAN-HD~\citep{tian2021good}      
    & 1729.6 & 2606.5 & 164.1 & 878.1 \\
    DIGAN~\citep{yu2022digan}           
    & 1630.2 & 2293.7 & 83.11 & 196.7 \\
    StyleGAN-V~\citep{skorokhodov2021stylegan}      
    & 1431.0 & 1773.4 & 79.52 & 197.0  \\
\midrule
    \rowcolor{aliceblue}
    \rowcolor{aliceblue}
    \sname-S (ours); 100/20-s   & \phantom{0}{457.4}  & \phantom{0}{902.2} & {71.46} & {159.9}\\
    \rowcolor{aliceblue}
    \sname-L (ours); 200/200-s   & \phantom{0}{398.9} & \phantom{0}\textbf{639.7} & {61.70} & {137.2} \\
    \rowcolor{darkblue}
    \sname-L (ours); 400/400-s   & \phantom{0}\textbf{343.6} & \phantom{0}{648.4} & \textbf{55.41}  & \textbf{125.2} \\
    \bottomrule
\end{tabular}
\end{minipage}
~~~
\begin{minipage}[hbt!]{.32\textwidth}
\captionof{table}
{IS values (higher values are better) of video generation models on UCF-101. Bolds indicate the best results and subscripts denote the standard deviations. 
* denotes the model is trained on train+test split, otherwise the method uses only the train split for training.
}\label{tab:is} 
\centering\small
\begin{tabular}{lc}
    \toprule
    Method & IS $\uparrow$ \\
    \midrule
    MoCoGAN~\citep{tulyakov2018mocogan} & 12.42\stdv{0.07}  \\
    ProgressiveVGAN~\citep{acharya2018towards} & 14.56\stdv{0.05}  \\
    LDVD-GAN~\citep{kahembwe2020lower}  & 22.91\stdv{0.19}  \\
    VideoGPT~\citep{yan2021videogpt}    & 24.69\stdv{0.30}  \\
    TGANv2~\citep{saito2020train}       & 28.87\stdv{0.67}  \\
    StyleGAN-V*~\citep{skorokhodov2021stylegan} & 23.94\stdv{0.73} \\ 
    DIGAN~\citep{yu2022digan}           & 29.71\stdv{0.53}  \\
    VDM*~\citep{ho2022video}            & 57.00\stdv{0.62}  \\
    TATS~\citep{ge2022long}             & 57.63\stdv{0.24}  \\
    \midrule
    \rowcolor{darkblue}
    \sname-L (ours)   &  \textbf{74.40\stdv{1.25}} \\
    \bottomrule
\end{tabular}
\end{minipage}
\vspace{-0.1in}
\end{figure*}

\vspace{0.02in}
\noindent\textbf{Evaluation.}
For evaluation metrics for quantitative comparison, we use and report Inception score (IS) \citep{salimans2016improved} and Fr\'echet video distance (FVD) \citep{unterthiner2018towards}. We use the clip length of 16 for the evaluation of IS, following the prior experiment setups in unconditional video generation. For FVD to evaluate UCF-101 and SkyTimelapse, we used a \emph{fixed protocol} proposed by StyleGAN-V that removes the potential bias from the existence of long videos in the dataset (we provide more details of metrics in Appendix~\ref{appen:metrics}). We consider two different clip lengths (16 and 128) for FVD, where we denote FVD$_{16}$ and FVD$_{128}$ as the FVD score measured on video clips of lengths 16 and 128, respectively. We use 2,048 real/fake video clips for evaluating FVD$_{16}$ and FVD$_{128}$, and generate 10,000 video clips for the evaluation of IS.

\noindent\textbf{Baselines.}
Following the setup in StyleGAN-V, one of the state-of-the-art video generation methods, we mainly compare \sname with the following recent video synthesis methods: VideoGPT~\citep{yan2021videogpt}, MoCoGAN~\citep{tulyakov2018mocogan}, MoCoGAN-HD \citep{tian2021good}, DIGAN~\citep{yu2022digan}, and StyleGAN-V~\citep{skorokhodov2021stylegan}. Moreover, we perform an additional comparison between \sname and previous approaches on IS values on UCF-101: 
MoCoGAN, ProgressiveVGAN~\citep{acharya2018towards}, LDVD-GAN~\citep{kahembwe2020lower}, VideoGPT, TGANv2~\citep{saito2020train}, DVD-GAN \citep{clark2019adversarial}, DIGAN, VDM~\citep{ho2022video}, and TATS~\citep{ge2022long}. All reported values are collected from the recent prior works: StyleGAN-V, DIGAN and TATS, unless otherwise specified. In particular, we compare the memory and computation efficiency with VDM. We provide more details of each method and how they are implemented in Appendix~\ref{appen:baselines}.

\vspace{0.02in}
\noindent\textbf{Training details.}
We use Adam~\citep{kingma2013auto} and AdamW optimizer~\citep{loshchilov2017decoupled} for training of an autoencoder and a diffusion model, respectively, where other training details mostly follow the setups in latent diffusion models for images~\citep{rombach2021highresolution}. For the autoencoder architecture, we use TimeSformer~\citep{bertasius2021space} to both encoder and decoder, where we use Transformer~\citep{vaswani2017attention} architecture for 3D-to-2D projection mapping. We use two configurations for diffusion models, (denoted by PVDM-S and PVDM-L, respectively); please refer to Appendix~\ref{appen:hyper} for more training details and model configurations.

\subsection{Main results}
\label{subsec:exp_main}

\noindent\textbf{Qualitative results.}
Figure~\ref{fig:main} illustrates the video synthesis results from \sname on UCF-101 and SkyTimelapse: our method shows realistic video generation results in both scenarios, namely, including the motion of the small object (Figure~\ref{subfig:ucf}) or the large transition of the over frames (Figure~\ref{subfig:sky}). We also note that such training is achieved under high-fidelity (256$\times$256 resolution) videos. Moreover, our method also has the capability to synthesize videos in the complex UCF-101 dataset with a plausible quality, as shown in Figure~\ref{subfig:ucf}, while other baselines often fail on such challenging dataset datasets~\citep{yu2022digan,skorokhodov2021stylegan}.\footnote{We provide the illustrations of synthesized videos from our method and comparison with other baselines in the following project website: \url{https://sihyun.me/PVDM}.}

\begin{table*}[t]
\centering
\begin{subfigure}{0.85\textwidth}
\centering
\includegraphics[width=\textwidth]{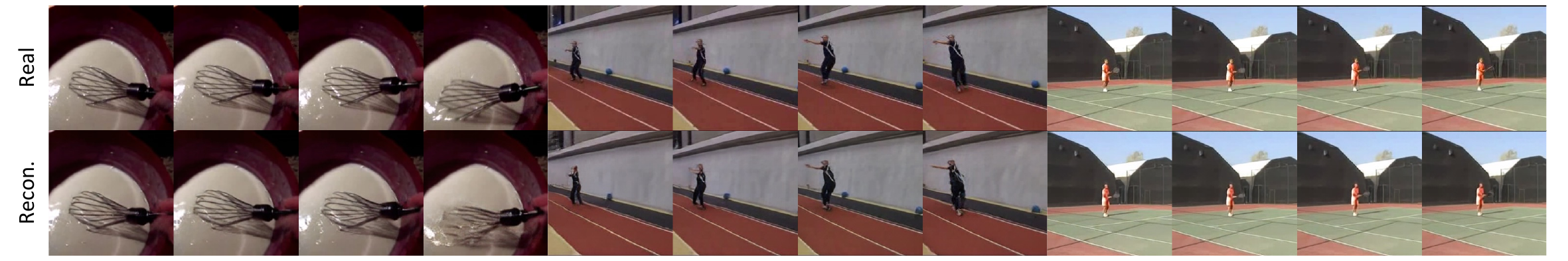}
\caption{UCF-101} 
\end{subfigure}
\begin{subfigure}{0.85\textwidth}
\centering
\includegraphics[width=\textwidth]{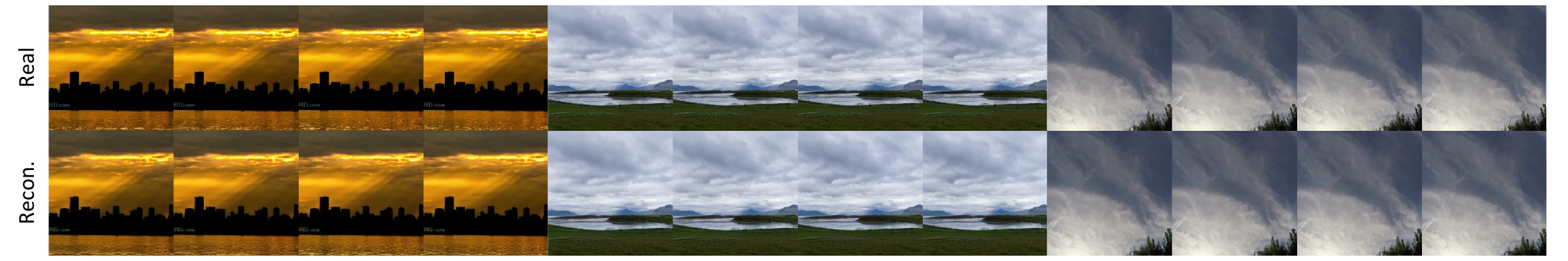}
\caption{SkyTimelapse} 
\end{subfigure}
\captionof{figure}{
Reconstruction results from our autoencoder trained on (a) UCF-101 and (b) SkyTimelapse. We visualize the frames with stride 4.
}
\vspace{-0.1in}
\label{fig:recon}
\end{table*}

\vspace{0.02in}
\noindent\textbf{Quantitative results.}
Table~\ref{tab:main} and \ref{tab:is} summarize the quantitative comparison between our method and prior video generation methods: \sname consistently outperforms other methods measured with diverse metrics. In particular, compared with VDM~\citep{ho2022video}, which trains a diffusion model on video pixels, our method shows a better IS even though VDM uses more data (uses train+test split) than our setup (uses train split only). Intriguingly, our method shows a notable improvement on UCF, a complicated, multi-class dataset, which shows the potential of our method to model complex video distribution. We also note that our method also achieves state-of-the-art results in 128 frame videos, which demonstrates the effectiveness of our method in generating longer videos. We provide the qualitative comparison with other video generation methods in Appedix~\ref{appen:quali}.

\vspace{0.02in}
\noindent\textbf{Long video generation.}
We visualize the long video generation results of our \sname in Figure~\ref{fig:emph}. As shown in this Figure, our \sname has the powerful capability of generating long videos (128 frames) with 256$\times$256 resolution frames while maintaining temporal coherency across timesteps. Here, we emphasize that our method produces long videos not only on fine-grained datasets with monotonic motion (\eg, SkyTimelapse) but also on UCF-101, which is a complex dataset that contains various dynamic motions. We also note that synthesizing long videos on such a complex UCF-101 dataset has been regarded as a challenging task in prior state-of-the-art video generation approaches that target long video generation~\citep{yu2022digan, skorokhodov2021stylegan, ge2022long}, and thus they often fail to generate temporally coherent and realistic videos on these datasets. We remark that the superiority of our method on long video synthesis is also verified quantitatively as $\text{FVD}_{128}$ in Table~\ref{tab:main}: our method significantly outperforms prior methods on this metric on UCF-101 as 1773.4 $\to$ 639.7.

\begin{table}[t]
\centering\small
\captionof{table}{Quantitative evaluation results between reconstruction from the autoencoder of \sname and the real videos. }
\label{tab:inter_extra}
\begin{tabular}{l c c c c c c c c}
\toprule
& \multicolumn{2}{c}{UCF-101} & \multicolumn{2}{c}{SkyTimelapse} 
\\
\cmidrule(lr){2-3} \cmidrule(lr){4-5} 
& Train & Test & Train & Test 
\\
\midrule
R-FVD $\downarrow$  
& 25.87 & 32.26 & \phantom{0}7.37 & \phantom{$^*$}36.52$^*$ 
\\
PSNR $\uparrow$     
& 27.34 & 26.99 &  34.33 & \phantom{$^*$}32.68$^*$
\\
\bottomrule
\end{tabular}
\\
\vspace{0.02in}
\scriptsize 
$*$ Evaluated with 196 samples due to the smaller test set size (196) than 2,048.
\vspace{-0.25in}
\end{table}

\subsection{Analysis}
\label{subsec:exp_ablation}
\noindent\textbf{Reconstruction quality.}
Figure~\ref{fig:recon} and Table~\ref{tab:inter_extra} summarize the results of the reconstructed videos from our autoencoder. We consider the following metrics to measure the quality: peak signal-to-noise-ratio (PSNR) for reconstruction quality and R-FVD for perceptual similarity. Here, R-FVD indicates FVD between reconstructions and the ground-truth real videos. Measured with these metrics, our method consistently shows accurate reconstructions similar to ground-truth videos. In particular, our method shows a small enough R-FVD, which validates the effectiveness of our latent space in compressing videos while preserving perceptual similarity. We also provide a more extensive analysis, including a comparison with other popular autoencoders in Appendix~\ref{appen:ae_abla}.

\vspace{0.02in}
\noindent\textbf{Comparison with VDM.}
Compared with a recent video diffusion model (VDM)~\citep{ho2022video}, our model achieves great computation and memory efficiency in generating samples due to the use of 2D instead of 3D convolution networks and the low-dimensionality of latent vector that encodes high-dimensional video pixels. Following the model configurations of VDM as described in~\cite{ho2022video}, Table~\ref{tab:memory} compares the maximum batch size that can be allocated in training as well as the memory and time to synthesize a single video at a 256$\times$256 resolution. Under the same sampler setup, PVDM achieves $\approx$17.6$\times$ better computation efficiency: our method requires $\approx$7.88 seconds to generate a video with 256$\times$256 resolution and length of 16, while VDM requires $>$2 minutes to generate such videos. Moreover, our PVDM shows at most 3.5$\times$ better memory efficiency; it can be trained with videos at 256$\times$256 resolution and length of 16 and synthesize longer videos (\eg, length 128) under the limited memory constraint (24GB), yet VDM cannot be trained and generate these videos under these computational resources. We also provide the comparison with two recent autoregressive video synthesis methods (TATS and VideoGPT): PVDM still achieves superior efficiencies in both time and memory.

\begin{table}[t]
\centering\small
\captionof{table}{Maximum batch size for training and time (s), memory (GB) for synthesizing a $256\times256$ resolution video measured with a single NVIDIA 3090Ti 24GB GPU. N/A denotes the values cannot be measured due to the out-of-memory problem. $N$/$M$-s denotes the model is evaluated with the DDIM sampler~\citep{song2021denoising} with $N$ steps (for the initial clip) and $M$ steps (for future clips).}
\vspace{-0.05in}
\resizebox{0.9\linewidth}{!}{
\begin{tabular}{lccc}
\toprule
& \multicolumn{1}{c}{Train}  & \multicolumn{2}{c}{Inference (time/memory)}  \\
\cmidrule(lr){2-2} \cmidrule(lr){3-4}
Length $\rightarrow$ & 16 & 16 & 128 \\
\midrule
TATS~\citep{ge2022long}
& 0 & 84.8/18.7 & 434/19.2\\
VideoGPT~\citep{yan2021videogpt} 
& 0 & \phantom{.}139/15.2  & N/A\\
\midrule
VDM~\citep{ho2022video}; 100/20-s     
& 0 & \phantom{.}113/11.1  & N/A \\
\rowcolor{aliceblue}
\sname-L~(ours); 200/200-s               
& 2 & 20.4/5.22  & \phantom{.}166/5.22  \\
\rowcolor{aliceblue}
\sname-L~(ours); 400/400-s               
& 2 & 40.9/5.22  & \phantom{.}328/5.22  \\
\rowcolor{aliceblue}
\sname-S~(ours); 100/20-s              
& \textbf{7} & \textbf{7.88}/\textbf{4.33} & \textbf{31.3}/\textbf{4.33} \\
\bottomrule
\end{tabular}
}
\label{tab:memory}
\vspace{-0.18in}
\end{table}

\vspace{-0.01in}
\section{Conclusion}
\vspace{-0.01in}
We proposed \sname, a latent diffusion model for video generation. Our key idea is based on proposing an image-like 2D latent space that effectively parameterizes a given video so that the given video data distribution can be effectively learned via diffusion models in latent space. 
We hope our method will initiate lots of intriguing directions in effectively scaling video synthesis methods.

\vspace{-0.05in}
\section*{Acknowledgements}
\vspace{-0.05in}
We thank Jaehyung Kim, Jihoon Tack, Younggyo Seo, and ‪Mostafa Dehghani‬ for their helpful discussions, and Lu Jiang and David Salesin for the proofreading our manuscript. We also appreciate Ivan Skorokhodov for providing baseline results and pre-trained checkpoints. This work was mainly supported by Institute of Information \& communications Technology Planning \& Evaluation (IITP) grant funded by the Korea government (MSIT) (No.2021-0-02068, Artificial Intelligence Innovation Hub; No.2019-0-00075, Artificial Intelligence Graduate School Program (KAIST)). This work is in part supported by Google Research grant and Google Cloud Research Credits program.

{\small
\bibliographystyle{ieee_fullname}
\bibliography{references}
}

\clearpage
\appendix
\onecolumn
\begin{center}
{\bf {\Large Appendix: Video Probabilistic Diffusion Models in Projected Latent Space}} 
\end{center}
\begin{center}
    \textbf{Project page:} \url{https://sihyun.me/PVDM}
\end{center}

\section{Detailed description of training objective}
\label{appen:objective}
In what follows, we describe the training objective that we used for our autoencoder consists of the encoder $f_{\bm{\phi}}$ and the decoder $g_{\bm{\psi}}$. We use the similar objective $\mathcal{L}(\bm{\phi}, \bm{\psi})$ used in VQGAN~\citep{rombach2021highresolution} and TATS~\citep{ge2022long}, except using vector-quantization regularization:
\begin{align*}
    \mathcal{L}(\bm{\phi}, \bm{\psi}) \coloneqq \mathcal{L}_{\tt pixel}(\bm{\phi}, \bm{\psi}) + \lambda_1 \mathcal{L}_{\tt LPIPS}(\bm{\phi}, \bm{\psi}) + \lambda_2\max_{h} \mathcal{L}_{\tt GAN} (\bm{\phi}, \bm{\psi})
\end{align*}
where $\mathcal{L}_{\tt pixel}$, $\mathcal{L}_{\tt LPIPS}$, $\mathcal{L}_{\tt GAN}$, $\mathcal{L}_{\tt VQ}(\bm{\phi},\bm{\psi})$ denote pixel-level $\ell_1$-reconstruction loss, negative perceptual similarity~\citep{zhang2018perceptual}, and adversarial objective~\citep{goodfellow2014generative} with a discriminator network $h$, respectively. Unlike prior works that uses KL-regularization~\citep{rombach2021highresolution} or VQ-regularization~\citep{ge2022long} to control the latent, we find simply adding $\tanh$ activation after projection network works well in training latent diffusion without without degradation autoencoder training.

For $\mathcal{L}_{\tt GAN}$, we use the sum of hinge-loss and feature similarity between real samples and their reconstructions, following the objective used in TATS to train the autoencoder. 
Following the training setup in VQGAN, we choose $\lambda_1=1$ and $\lambda_3=0.25$ as a constant and set $\lambda_2=0$ before the other terms converge and $\lambda_2=0.25$ after the convergence. We find early stopping is crucial in training our autoencoder with $\mathcal{L}_{\tt GAN}$; we stop the training by tracking the R-FVD values during training.

\section{Detailed description of experiment setup}
\subsection{Datasets}
\noindent\textbf{UCF-101} \citep{soomro2012ucf101} is a video dataset composed of diverse human actions. Specifically, the dataset contains 101 classes, and each video has a 320$\times$240 resolution. There are 13,320 videos in total, where we used only the train split (9,357 videos) for training and the rest of the test split for the evaluation, following the common practice in evaluating unconditional video generation on UCF-101~\citep{yu2022digan}. All videos are center-cropped and resized to $256\times256$ resolution.

\vspace{0.02in}
\noindent\textbf{SkyTimelapse} \citep{xiong2018learning} is a collection of sky time-lapse total of 5,000 videos. We use the same data preprocessing following the official link.\footnote{\url{https://github.com/weixiong-ur/mdgan}} Moreover, similar to UCF-101, all videos are center-cropped and resized to $256\times256$ resolution. We use the train split for both training the model and the evaluation, following the evaluation setup used in StyleGAN-V~\citep{skorokhodov2021stylegan}.

\label{appen:dataset}

\subsection{Metrics}
\label{appen:metrics}
For a fair comparison with prior works, we carefully choose the setups for quantitative evaluation. For measuring the Inception score (IS; \citet{salimans2016improved}) on UCF-101~\citep{soomro2012ucf101}, we use the C3D network~\citep{tran2015learning} trained on the Sports-1M dataset~\citep{karpathy2014large} and fine-tuned on UCF-101. Following the representative setups for evaluating IS for videos, also used in recent state-of-the-art methods~\citep{yu2022digan, ge2022long}, we generate 10,000 samples to measure the score.

In the case of  Fr\'echet video distance (FVD; \citet{unterthiner2018towards}), we used the fixed protocol proposed by StyleGAN-V~\citep{skorokhodov2021stylegan}. Unlike predominant setups that first preprocess the given video dataset as a fixed-length video clip and then sample real samples for calculating real statistics, the paper proposes first to sample the video data and randomly choose the fixed-length video clip. Such a different protocol is inspired by the observation from the previous evaluation procedure---if the dataset contains extremely long videos, the statistics can be easily biased due to the large number of clips from this video. We sample 2,048 samples (or the size of the real data if it is smaller) for calculating real statistics and 2,048 samples for evaluating fake statistics.

\clearpage
\section{Detailed description of the baselines}
\label{appen:baselines}
In this section, we describe the main idea of video generation methods that we used for the evaluation.

\begin{itemize}[leftmargin=0.2in]
\item \textbf{VGAN} \citep{vondrick2016generating} 
replaces 2D convolutional networks to 3D convolutional networks to extend image generative adversarial network (GAN; \citep{goodfellow2014generative}) architecture for video generation.

\item \textbf{TGAN} \citep{saito2017temporal} extends Wasserstein GAN \citep{arjovsky2017wasserstein} for images based on separating the motion and the content generator.

\item \textbf{MoCoGAN} \citep{tulyakov2018mocogan} decomposes motion and the content of the video for video generation via having a separate content generator and an autoregressive motion generator.

\item \textbf{ProgressiveVGAN} \citep{acharya2018towards} Inspired by GAN-based high-resolution image generation via progressive architecture design~\citep{karras2017progressive}, ProgressiveGAN also generates the video progressively in both spatial and temporal directions.

\item \textbf{LDVD-GAN} \citep{kahembwe2020lower} mitigates computation inefficiency of video GANs with a discriminator using low-dimensional kernels.

\item \textbf{VideoGPT} \citep{yan2021videogpt} is a two-stage model: it encodes videos as a sequence of discrete latent vectors using VQ-VAE~\citep{van2017neural} and learns the autoregressive model with these sequences via Transformer~\citep{vaswani2017attention}.

\item \textbf{TGANv2} \citep{saito2020train} proposes a computation-efficient video GAN based on designing submodules for a generator and a discriminator.

\item \textbf{DVD-GAN} \citep{clark2019adversarial} uses two discriminators for identifying real/fake of each spatial and temporal dimension, where the input of the temporal discriminator is low-resolution for efficiency.

\item \textbf{MoCoGAN-HD} \citep{tian2021good} proposes to use a strong image generator for high-resolution image synthesis: they generate videos via modeling trajectories in the latent space of the generator. 

\item \textbf{DIGAN}~\citep{yu2022digan} proposes a video GAN based on exploiting the concept of implicit neural representations and computation-efficient discriminators. 

\item \textbf{StyleGAN-V}~\citep{skorokhodov2021stylegan} also introduces neural-representation-based video GAN to learn a given video distribution with a computation-efficient discriminator.

\item \textbf{TATS}~\citep{ge2022long} proposes a new VQGAN~\citep{esser2020taming} for videos and trains an autoregressive model to learn the latent distribution.

\item \textbf{VDM}~\citep{ho2022video} extends image diffusion models by proposing a new U-Net architecture based on 3D convolutional layers.

\end{itemize}

\clearpage
\section{More details on training setup}
\label{appen:hyper}
For all experiments, we use a batch size of 24 and a learning rate 1$e$-4 for training autoencoders. We train the model until both FVD and PSNR converge. We use 4-layer Transformer for 3D-to-2D projections, where the number of heads is set to 4, the hidden dimension is set to 384, and the hidden dimension of the multilayer perceptron (MLP) in Transformer is to be 512. We set the dimension of the latent codebook to 4. For training diffusion models, we set a learning rate as 1$e$-4 with a batch size of 64. For more hyperparameters, such as the base channel and the depth of U-Net architecture, we adopt a similar setup to LDMs~\citep{rombach2021highresolution}. In particular, we set the codebook channel $C$ to 4 and the patch size to $4\times4\times1$, so that a $256\times256\times16 \times 3$ dimension video is encoded to a $(32\times32 + 32\times16 + 32\times16) \times4 =$ $8$,$192$ dimension vector. Throughout the experiments, we consider two model configurations: PVDM-S (small model) and PVDM-L (large model), where we summarize the detailed configurations and hyperparameters in Table~\ref{tab:config}. 
\begin{table}[h]
    \centering
    \resizebox{\linewidth}{!}{
    \begin{tabular}{l ccccccccc}
    \toprule
         & {Base channel} & {Attn. res.} & {\# ResBlock} & {Channel mul.} & {\# heads} & {Linear start} & {Linear end} & {Timesteps} & {\# Iter.} \\
         \midrule
         {PVDM-S} & 128 & {[4,2,1]} & 2 & {[1,2,4]} & 8 & 0.0015 & 0.0195 & 1000 & 400k \\
         {PVDM-L} & 256 & {[4,2,1]} & 2 & {[1,2,4]} & 8 & 0.0015 & 0.0195 & 1000 & 850k \\
    \bottomrule
    \end{tabular}
    }
    \caption{Model configurations for PVDM-S and PVDM-L.}
    \label{tab:config}
\end{table}

\section{Qualitative comparison of generated results}
\label{appen:quali}

\begin{figure*}[h]
\begin{subfigure}{0.49\textwidth}
\centering
\includegraphics[width=\textwidth]{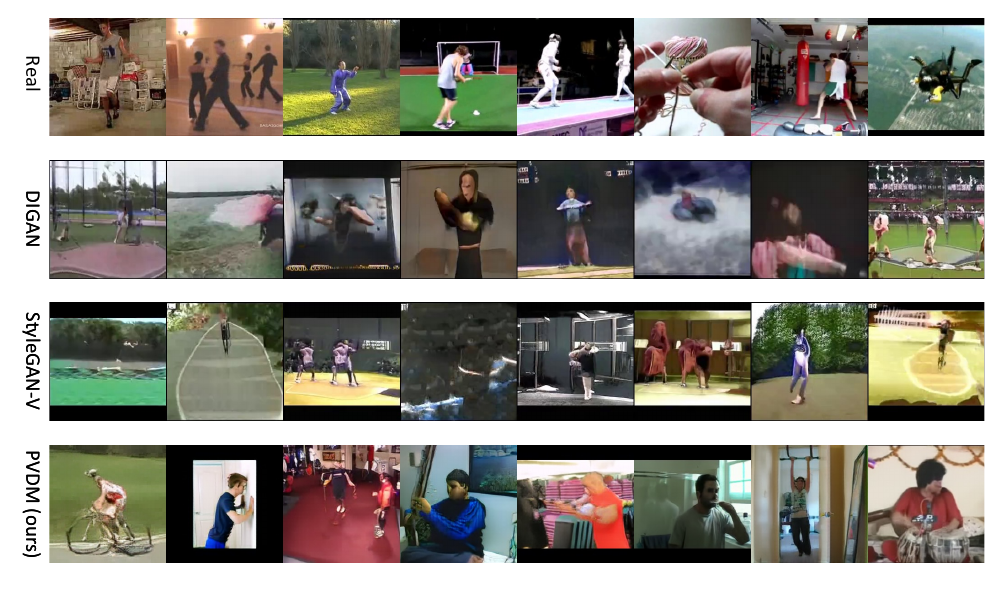}
\caption{UCF-101} 
\end{subfigure}
\hfill
\begin{subfigure}{0.49\textwidth}
\centering
\includegraphics[width=\textwidth]{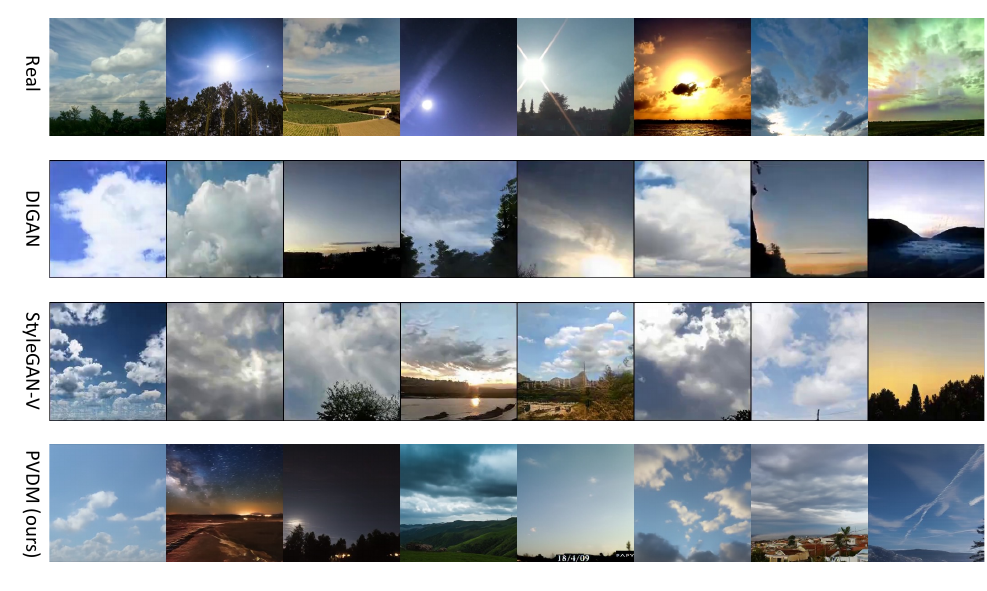}
\caption{SkyTimelapse} 
\end{subfigure}
\caption{
Illustrations of the synthesized first frame results of \sname and baselines trained on UCF-101 and SkyTimelapse datasets. Baseline results are from the StyleGAN-V website:~\url{https://universome.github.io/stylegan-v}.
}
\label{fig:quali_compare}
\end{figure*}

We provide the qualitative comparison of the first frame generated from baselines in Figure~\ref{fig:quali_compare}; note that we provide the comparison of videos on our anonymized website. Compared with other recent video synthesis methods, our \sname shows overall high-quality synthesis results, especially on the complex dataset, UCF-101. 

\clearpage
\section{More analysis in our autoencoder architecture}
\label{appen:ae_abla}
\vspace{0.02in}
\noindent\textbf{Ablation study on our autoencoder.}
We also conduct an ablation study of autoencoder on UCF-101 and report it in Table~\ref{tab:ae_abla} (we adjust the model sizes to be equal): Our autoencoder components are particularly effective in capturing perceptual details with ``2D-shaped'' latents $\mathbf{z}$, as shown by better R-FVD than other variants. We also find the performance of 2D autoencoder degrades a lot for small $\dim(\mathbf{z})$ (8,192) as 2D AE has no temporal layers and thus is hard to capture the temporal coherence with compact latents. We also find removing popular vector quantization used for training autoencoders is not necessary in improving our autoencoder quality and does not prevents diffusion model training; it was enough to clip the range of latents by adding activation functions after the encoder ouputs (e.g., $\mathtt{tanh}$).

\vspace{0.02in}
\noindent\textbf{Autoencoder performance with various model sizes.}
Table~\ref{tab:ae_size} summarizes the autoencoder performance \wrt varying model sizes.
It shows our autoencoder is scalable and it works even fairly well using $\approx\nicefrac{2}{3}$ number of parameters.

\begin{figure*}[t]
\begin{minipage}[hbt!]{.64\textwidth}
\centering\small
\captionof{table}{Ablation study of our projected autoencoder.}
\begin{tabular}{l c c c c c c}
\toprule
Backbone  & Proj. & VQ. & $\text{dim}(\mathbf{z})$ & $\mathbf{z}$ shape & PSNR $\uparrow$ & R-FVD $\downarrow$ 
\\
\midrule
{2D CNN} & - & \checkmark
& 32,768 & 3D & 25.23 & 147.5 \\
\midrule
2D CNN & - & \checkmark
& \phantom{0}8,192& 3D & 21.89 & 559.9 \\
{Timesformer} & - & \checkmark
& \phantom{0}8,192 & 3D  &24.65&  134.9 \\   
{Timesformer} & \checkmark & \checkmark    
& {\phantom{0}8,192} & {2D} & {24.73}& {63.34}  \\ 
\rowcolor{darkblue}
{Timesformer} & \checkmark & -
& {\phantom{0}8,192} & {2D} & {26.99}& {32.26}  \\ 
\bottomrule
\end{tabular}
\label{tab:ae_abla}
\end{minipage}
\begin{minipage}[hbt!]{.32\textwidth}    
\centering\small
\captionof{table}{Autoencoder performance (R-FVD) with different model sizes. \# Enc. and \# Dec. denote the number of parameters for encoder and decoder, respectively.}
\begin{tabular}{c c}
    \toprule
    \# Enc.+\# Dec. & R-FVD $\downarrow$ 
    \\
    \midrule
    11M+10M & 154.7 \\   
    23M+17M & 67.76 \\
    35M+27M & 63.34 \\ 
    \bottomrule
    \end{tabular}
    \label{tab:ae_size}
\end{minipage}
\end{figure*}

\section{Discussion with concurrent work}
The concurrent work, MagicVideo~\citep{zhou2022magicvideo}, is also a latent video diffusion model. However, unlike our PVDM, they encode videos in a frame-wise manner, resulting in an unfavorable increase of latent dimension with length and thus limiting the scalability. Another concurrent work, LVDM~\citep{he2022lvdm} uses 3D CNN to encode video clip and design a latent diffusion model, but they still deal with cubic-shaped latent tensors in designing diffusion models. Make-a-video~\citep{singer2022make} extends a popular image diffusion model for video synthesis. They also avoid using 3D convolutions, yet they deal with 3D RGB arrays directly and are still difficult to be trained on high-resolution, long videos. As a result, it leads the framework to heavily rely on complex spatiotemporal interpolation modules to generate high-dimensional videos.

\section{Limitations and future work}
While \sname shows promise in efficient and effective video synthesis, there are many possible future directions for further improvement as there still exists a gap between the real and the generated videos. For instance, designing diffusion model architectures specialized for our triplane latents to model latent distribution better or investigating better latent structures than our triplane idea to encode video more efficiently would be interesting directions. While we have not conducted experiments in large-scale video datasets to perform challenging text-to-video generation due to the limited resources, we strongly believe our framework also works well in this task and leave it for the future work.

\end{document}